\documentclass[10pt,letterpaper]{article}
\usepackage[top=0.85in,left=2.75in,footskip=0.75in]{geometry}

\usepackage{amsmath,amssymb}

\usepackage{changepage}

\usepackage[utf8x]{inputenc}

\usepackage{textcomp,marvosym}

\usepackage{cite}

\usepackage{nameref,hyperref}

\usepackage[right]{lineno}

\usepackage{microtype}
\DisableLigatures[f]{encoding = *, family = * }

\usepackage[table]{xcolor}

\usepackage{array}

\usepackage{booktabs}
\usepackage{multirow}
\usepackage{lscape}	
\usepackage{tabulary}
\usepackage{graphicx}
\usepackage{enumitem}
\usepackage{makecell}
\usepackage{epstopdf} 
\usepackage{dsfont}
\usepackage[linesnumbered,ruled,vlined]{algorithm2e}
\usepackage{multicol}
\usepackage[export]{adjustbox}
\usepackage{float}
\usepackage{algorithm2e,etoolbox}
\usepackage{anyfontsize}
\usepackage{verbatim} 
\AtBeginEnvironment{algorithm}{}

\newcolumntype{+}{!{\vrule width 2pt}}

\newlength\savedwidth

\raggedright
\usepackage[aboveskip=1pt,labelfont=bf,labelsep=period,justification=raggedright,singlelinecheck=off]{caption}

\makeatletter
\renewcommand{\@biblabel}[1]{\quad#1.}
\makeatother

\usepackage{lastpage,fancyhdr,graphicx}
\usepackage{epstopdf}
\fancyheadoffset[L]{2.25in}
\fancyfootoffset[L]{2.25in}
\begin{document}
\vspace*{0.2in}

\begin{flushleft}
{\Large
\textbf\newline{DTI-SNNFRA: Drug-Target interaction prediction by shared nearest neighbors and fuzzy-rough approximation} 
}
\newline
\\
Sk Mazharul Islam\textsuperscript{1},
Sk Md Mosaddek Hossain\textsuperscript{2*},
Sumanta Ray\textsuperscript{2}
\\
\bigskip
\textbf{1} Department of Computer Science and Engineering, RCC Institute of Information Technology, Kolkata, West Bengal, India
\\
\textbf{2} Department of Computer Science and Engineering, Aliah University, Kolkata, West Bengal, India
\\
\bigskip

%
%





* mosaddek.hossain@gmail.com

\end{flushleft}
\section*{Abstract}
\emph{In-silico} prediction of repurposable drugs is an effective drug discovery strategy that supplements \textit{de-nevo} drug discovery from scratch. Reduced development time, less cost and absence of severe side effects are significant advantages of using drug repositioning. Most recent and most advanced artificial intelligence (AI) approaches have boosted drug repurposing in terms of throughput and accuracy enormously. However, with the growing number of drugs, targets and their massive interactions produce imbalanced data which may not be suitable as input to the classification model directly. Here, we have proposed DTI-SNNFRA, a framework for predicting drug-target interaction (DTI), based on shared nearest neighbour (SNN) and fuzzy-rough approximation (FRA). It uses sampling techniques to collectively reduce the vast search space covering the available drugs, targets and millions of interactions between them. DTI-SNNFRA operates in two stages: first, it uses SNN followed by a partitioning clustering for sampling the search space. Next, it computes the degree of fuzzy-rough approximations and proper degree threshold selection for the negative samples' undersampling from all possible interaction pairs between drugs and targets obtained in the first stage. Finally, classification is performed using the positive and selected negative samples. We have evaluated the efficacy of DTI-SNNFRA using AUC (Area under ROC Curve), Geometric Mean, and F1 Score. The model performs exceptionally well with a high prediction score of 0.95 for ROC-AUC. The predicted drug-target interactions are validated through an existing drug-target database (Connectivity Map (Cmap)).

\section{Introduction}
Drug development strategies, also known as drug repositioning or drug repurposing or drug reprofiling, predict the interaction among drugs and targets from the existing drug-target databases \cite{SACHDEV2019103159}. There are two types of drug-target interaction: competitive inhibitors and allosteric inhibitors. Competitive inhibitors adhere to the target's active site to suppress reactions. Allosteric inhibitors bind to the target's allosteric site, which in turn prevents reactions, correct metabolic imbalance, and kills pathogens to cure diseases. There exist several synthesized compounds whose target profiles and effects are still unknown. The research and findings of compounds' properties, their reactions/responses to drugs, and targets have generated large, complex databases that need efficient computational methods to analyze and predict drug-target interaction. New drug design requires more than 13.5 years and the cost exceeds 1.8 billion dollars \cite{Cui2019}, \cite{Ezzat2016}.
Moreover, new drugs may have unwanted side effects on patients. Therefore, due to known side effects and easier government approval, drug-repurposing facilitate pharmaceutical companies to launch existing authorized drugs and compounds in the market for new therapeutic purposes \cite{Bagherian2020}. Drug repositioning usually reinvestigates existing drugs which were denied approval due to new therapeutic indications. 

Practical laboratory experiments to discover the interactions among the drugs and targets are expensive, time-consuming and labour-intensive. Therefore, in-silico approaches are gaining attention, in which virtual screening is initially accomplished, and then possible candidates go through experimental verification. Docking simulations is a type of in-silico approach that need 3D structure analysis of drugs and target molecules to determine the potential binding sites. Despite the excellent accuracy of this process, unavailability of the proper 3D structure of drugs and targets, and long processing time hinders the docking simulation. Chemogenomics was introduced to tackle this problem in which the chemical space and genomic space are mined together to find the potential compounds such as imaging probes and drug leads \cite{Bagherian2020}. Plenty of machine learning techniques based on similarity computation, matrix factorization, network models, features vectors, and deep learning models for DTIs prediction are prevalent in the literature \cite{SACHDEV2019103159,DSOUZA2020748}. Similarity-based approaches find how a new drug and target is similar to known drug-target pairs based on the pharmacological similarities between drugs and the genomic similarity of protein sequences. Here, similarity measures may be either chemical-based, ligand-based, expression-based, side effect-based, or annotation-based \cite{Bagherian2020}. But the disadvantage of this approach is that only a tiny proportion of drug-target interaction pairs are known and available for comparison. There are many matrix factorization algorithms, in which given an interaction matrix $X_{n \times m}$, the main goal is to decompose it into two lower-order matrices, $Y_{n \times k}$ and $Z_{m \times k}$ such that $X=YZ^T$ with $k < n, m$ \cite{Bagherian2020}. The matrix completion technique is then used to compute the missing data that help in the DTI prediction task. In feature-based \cite{Bagherian2020} methods, the drug and target vector are concatenated. A binary or real label is then appended that denotes interaction outcome or affinity score for each drug-target pairs. Examples of features-based methods include the Bagging-based Ensemble method(BE-DTI) that employs dimensionality reduction, and active learning \cite{Sharma2018}.
In \cite{Ezzat2017}, first feature sub-spacing and then three different dimensionality reduction techniques, namely Singular Value Decomposition(SVD), Partial Least Squares (PLS), and Laplacian Eigenmaps (LapEig) are used to prepare training data. They have used decision tree and kernel ridge regression classifiers as base learners. Network-based models such as TL-HGBI, DrugNet utilizes heterogeneous networks not only to predict the drug but also recommend the way of treatments \cite{Bagherian2020}, \cite{Cui2019}. In \cite{Seal2015}, the matrix inverse computation is used to compute relevance grade between two nodes in a weighted network of drug-target interactions. Deep learning-based DTIs prediction utilizes the biological, topological, and physicochemical information of the drugs and targets to compute feature vectors/matrix \cite{Bagherian2020}, \cite{CHEN20181241}. They can capture the inherent drug-target interactions over other state-of-the-art feature computation methods and classifiers. Deep learning techniques sometimes can not be applied due to the unavailability of sufficient data.

In this article, a feature-based method, DTI-SNNFRA, is proposed. 
Here, we have represented each drug or target by a feature vector. Initially, all the approved drug-target pairs are considered as a set of positive samples. The remaining unannotated and non-approved interaction pairs from which interaction may be predicted can be initially treated as a set of negative samples. Here, the drug-target interaction prediction task is a class imbalance problem, as most interaction pairs are unannotated. 
Our proposed framework predicts DTI in two phases that considerably reduce the unannotated drug-target pairs' search space. In the first phase, from each known drug-target interaction pair, the shared nearest-neighbours (SNN) of the associated drug and target are computed using their feature vectors. Then, SNNs of the drug are clustered, and each cluster's centroid is taken as a representative. Representative targets are also determined similarly. These representative drugs and targets are used to form drug-target pairs that are fewer and are probable candidates for possible interactions. The pairs obtained in this way are treated as negative interaction pairs.

Despite the reduction in search space, the obtained training set created in this way is highly imbalanced. To encounter this problem, in the second phase, our prediction model computes a fuzzy rough upper approximation score (grade membership degree) as the strength of the interaction between a drug and target for each of the remaining unannotated pairs.
Based on this score's different threshold cut-off values, we have initially divided all the unannotated drug-target pairs into positive and negative classes. The size of the so obtained negative samples is dependent on the threshold cut-off, and if it is significantly larger than the size of the positive samples, then the drug-target interaction training dataset remains imbalanced.
On the other hand, if the number of unannotated negative samples is considerably less than the approved positive samples, oversampling is carried out by an Adaptive Synthetic Sampling Method (ADASYN). It produces a reduced and balanced training set that can be used by any general classifier. We have applied several state-of-the-art classifiers such as SVM, decision tree, random forest, and RUSBoost to find predicted interactions' correctness.

In section \ref{MaterialsAndMethods} of this article, the datasets utilized in this work along with method and algorithms, is explained. In section \ref{FRUAsection}, a brief description and definition of the fuzzy-rough set based lower and upper approximation are outlined. In section \ref{MetResDis}, results and discussions are presented and finally section \ref{conclu} draws the conclusion.

\section{Materials and methods}\label{MaterialsAndMethods}
\begin{figure}[!htb]
	\begin{center}
		\includegraphics[scale=1.0]{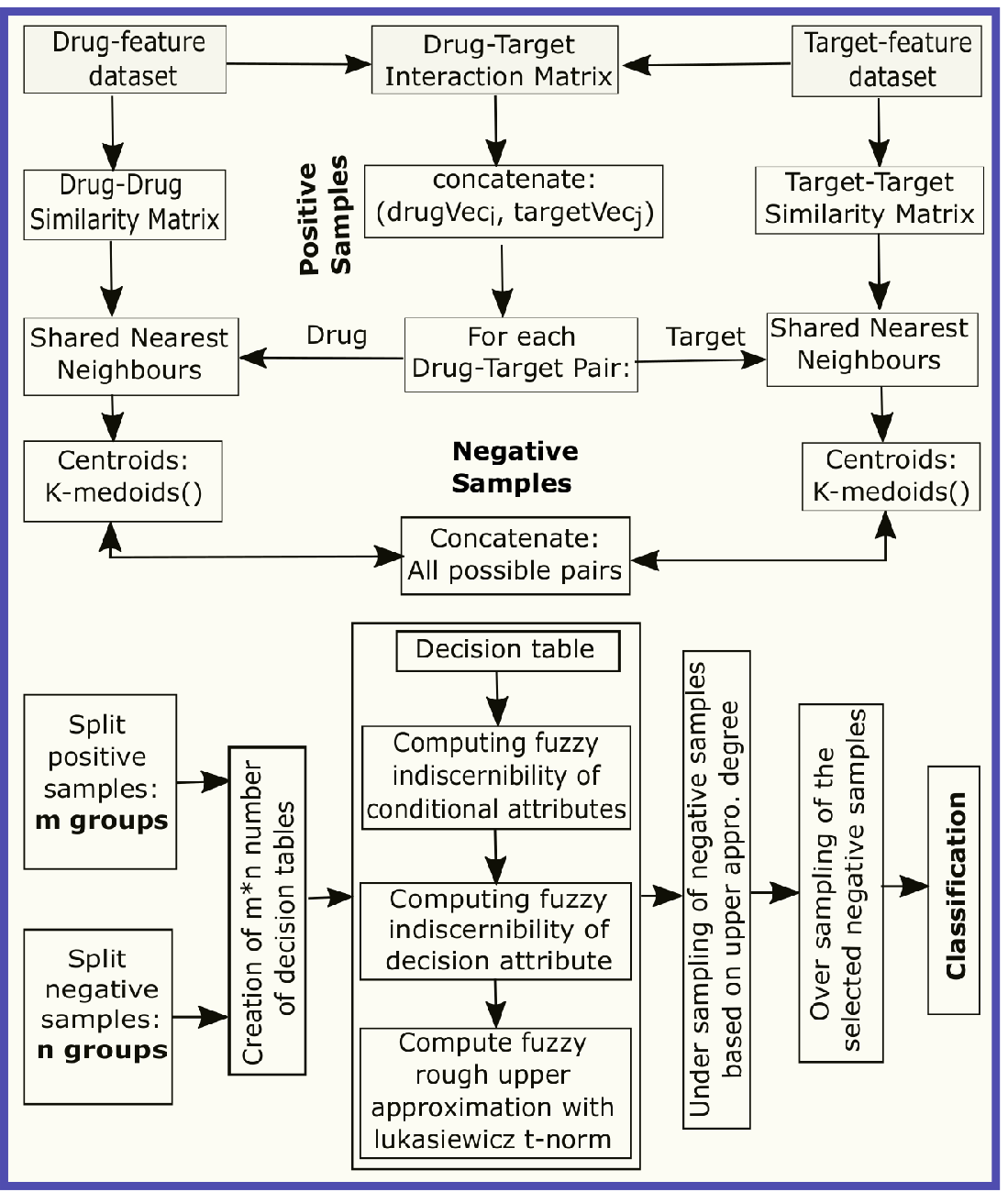}
		\caption{Building block of proposed DTI-SNNFRA Method}
		\label{Pro_1} 
	\end{center}
\end{figure}

In this section, we describe the datasets used in this work, key ideas of our algorithms, and some background of the fuzzy-rough set. The building block of the proposed DTI-SNNFRA method is shown in Figure \ref{Pro_1}.

\subsection{Dataset Preparation}
In this article, the drug-target interaction data is taken from the DrugBank database \cite{Knox2010} (version 4.3, released on 17 Nov. 2015) and from dataset mentioned in \cite{Tabei2012}. In dataset 1 \cite{Knox2010}, the number of drugs is 5877, targets are 3348, and the number of interactions between the drugs and targets is 12674. Here, a drug or a target is represented by its feature vector. The drug feature vector is computed by Rcpi \cite{Cao2014} package, and the PROFEAT \cite{Li2006} web server. It is represented by constitutional, topological, and geometrical descriptors. The target feature vector is computed using different types of compositions, such as amino acid, pseudo-amino acid, and CTD (composition, transition, distribution) descriptors. The number of features for drug and target of dataset 1 are 193 and 1290, respectively.

In dataset 2 \cite{Tabei2012}, the number of drugs is 1862, targets are 1554, and the number of interactions between the drugs and targets is 4809. Here, each drug is represented by a binary vector known as PubChem fingerprint. Each element of this vector exhibits the existence and non-existence of one of the 881 chemical substructures. Similarly, each target in the dataset 2 is also represented as a fingerprint of an 876-dimensional binary vector. Here, each element of this vector indicates the existence and non-existence of 876 different protein domains, as mentioned in the Pfam database \cite{PFAM2016}. The drug feature vector and target feature vector are then concatenated to represent the drug- target pair feature vector and can be represented for dataset 1 as:
\begin{center}
	$\{d_1, d_2,.....,d_{193},t_1,t_2,.....,t_{1290}\}$\\
\end{center}
These drug-target pairs feature vectors are then normalized in the range $[0, 1]$ by min-max method for avoiding bias towards any feature. 



\subsection{Workflow of the proposed framework}

In this section, the necessary steps of our proposed method are described.
\subsubsection{Step 1: Finding positive and negative drug-target pairs}\label{method_step1}

After the normalization, only the drugs and targets which have known interactions in the interaction matrix are used to form the positive samples for classifiers. But the number of unannotated and non-approved interaction pairs derived from the interaction matrix is significantly greater than the number of positive samples.
Note that the high dimensionality and numerous samples may have diverse effects in the prediction task. Finding characteristically similar drugs and targets using the nearest-neighbour search facilitates new drug-target prediction. Determination of the nearest-neighbours using similarity distance measures are sensitive to the dimensionality and the distribution of the dataset. The popular similarity function $L_1$ and $L_2$  in Minkowski space infers the fact that, for particular data distribution, if the dataset's dimensionality is increased then the relative difference of the distance of the closest and farthest data point of an independently selected point goes to $0$. For this reason, the primary distance functions like $L_1$, $L_2$, and cosine, etc. are not suitable for high dimensional data.
In this context, computing shared nearest neighbours (SNN) using the primary distance functions instead of computing nearest neighbours reduce the disadvantage of higher dimensions \cite{Houle2010}. Assume the dataset $S$ consisting of $n = |S|$ objects and $k \in N^{+}$. For each individual drug (or target), let $NN_k (x) \subseteq S$ represents $k$-nearest-neighbors of $x \in S$. It is computed using $L_2$ similarity measure. The overlap between the computed  $k$-nearest-neighbors sets of the objects $x$ and $y$ is represented as:

\begin{equation}
SNN_k (x, y) = |NN_k (x)\cap NN _k(y)|
\end{equation}
The Algorithm \ref{sharedNNAlgo}, provides the procedure to compute shared nearest neighbours, and the Algorithm \ref{DataSetPreparation}, outlines how the training dataset is prepared for classifiers.

Suppose, the total number of drugs and targets are $M$ and $N$. Assume drug $d_{i, i\in M}$  interacts with target $t_{j,j \in N}$. Now for this $d_i$, the indices of all drugs in $\bigcup SNN_k (d_i,d_r),\forall r\in M$ and $i\neq r$ are identified and assigned to $snnD_i$. Similarly, for the target $t_j$, the indices of all targets in $\bigcup SNN_k (t_j,t_r),\forall r\in N$ and $j\neq r$ are identified and assigned to $snnT_j$. Then all the drugs and targets in $snnD_i$ and $snnT_j$ are clustered using the k-medoids clustering and centroids are selected as a representatives of $snnD_i$ and $snnT_j$. The Calinski-Harabasz criterion is used here to determine the correct number of clusters. These representatives drugs and targets from $snnD_i$ and $snnT_j$ are used to construct cartesian product pairs. Subsequently, the corresponding drug vector and target vector are concatenated for each cartesian product pair, which are included in the negative samples set. Forming negative samples by the above SNN approach followed by k-medoids clustering reduces the inclusion of the irrelevant drug-target pairs. For example, in dataset 1, the number of approved drug-target pairs is $12674$, and the number of all possible pairs from which interaction may be predicted is $19663522$. The number of drug-target pairs selected by the SNN followed by k-medoids clustering is $45933$, which indicates $427$ times samples removal.

\subsubsection{Step 2: Decision table preparation and average approximation degree computation}
The positive and negative sets of samples obtained in \ref{method_step1} are divided into $m$ and $n$ groups, respectively. Each group from the negative set, say, $n_j$ is taken $m$ times with $m$ group from the positive set, and $m$ number of the decision table is prepared. Each decision table is used to compute the fuzzy rough upper approximation degree of each sample in the $n_j$ group. The $m$ number of upper approximation degree of each sample in the $n_j$ group are then taken for average upper approximation degree computation. In Algorithm \ref{FRUAAlgoPseudoCode}, We have mentioned this average upper approximation degree computation.

\subsubsection{Step 3: Under-sampling based on approximation degree}
A fuzzy rough grade membership is computed for every negative sample using all positive samples' interactions via Algorithm \ref{FRUAAlgoPseudoCode}. This fuzzy-rough upper approximation degree possibly indicates the possible interaction degree value between $0$ to $1$ scale. Now, one threshold value near $1$ called $th1$ can be assumed to select many samples whose fuzzy-rough upper approximation degree is smaller than or equal to $th1$.  Another one threshold value near $0$ called $th0$ can be assumed to select many samples whose fuzzy-rough upper approximation degree are less than or equal to $th0$. This $th0$  and $th1$ based sample selection both under-samples the negative samples set.

\subsubsection{Step 4: Oversampling, if required} \label{method_step4}

The datasets used here has several approved drug-target pairs, which are treated as a set of positive samples. The remaining pairs that are unannotated may or may not interact with each other. These unannotated (and also non-approved) interaction pairs are enormous, from which interaction is predicted. For example, in dataset 1, the number of approved drug-target pairs is 12674, and the number of remaining unannotated pairs is 19663522. Initially, we have reduced the number of unannotated pairs (i.e. initially treated as a set of negative samples), by using Shared Nearest Neighbor in Step \ref{method_step1}. The number of unannotated negative samples, previously selected by SNN, remains higher than positive samples. Our prediction model then computes a fuzzy rough upper approximation score (grade membership degree) as the strength of the interaction between a drug and target for each of the remaining unannotated pairs. Based on different threshold cut-off values of this score, we have initially divided all the unannotated drug-target pairs into positive and negative classes. The size of the so obtained negative samples is dependent on the threshold cut-off, and if it is significantly larger than the size of the positive samples, then the drug-target interaction training dataset remains imbalanced. Therefore, we have selected one threshold value of grade membership degree to under-sample the negative samples to get an approximately equal number of negative and positive samples. In this case, no oversampling is needed. However, if we select different threshold values where the number of negative samples is less than the number of positive samples, the oversampling of negative samples is required to balance negative and positive samples.

\subsubsection{Step 5: Interaction prediction}
As obtained in section \ref{method_step4}, the dataset is then used to predict the negative set's drug-target interaction pairs.

\subsection{Fuzzy rough set}\label{FRUAsection}
Assume that the drug-target pairs obtained by the given interaction matrix and SNN-based initial filtering constitute a decision table called $\mathcal{IT}$. In this table, every row is denoted by $m$ numbers of features i.e. $C=\{f_i:1\leq i \leq m\}$ and one decision attribute $D=\{d\}$.
In this $\mathcal{IT}$,  how two objects are indiscernible is determined by calculating fuzzy indiscernibility relation (FIR). Subsequently, this indiscernibility relation is taken to determine fuzzy-rough lower and upper approximation. The fuzzy lower and upper approximations using fuzzy similarity relation (either fuzzy equivalence or tolerance relation), in pursuance of Radzikowska's model, to approximate a concept $Y$ are defined as
\cite{ref35:fuzzyDiscernibility:JensenShen}:

\begin{eqnarray}
\label{RadzikowskaLower}
\mu_{\underline{R_P}Y}(x) = \underset{y \in \mathcal{IT}}{inf} I(\mu_{R_P}(x,y),\mu_{Y}(y) )\\
\mu_{\overline{R_P}Y}(x) = \underset{y \in \mathcal{IT}}{sup} T(\mu_{R_P}(x,y),\mu_{Y}(y) )
\label{RadzikowskaUpper}
\end{eqnarray}
Here, in equations \ref{RadzikowskaLower} and \ref{RadzikowskaUpper}, $I$ indicates a fuzzy implicator, $T$ denotes a $t$-norm and $R_P$ is the fuzzy similarity relation computed by the features subset $P \subseteq C$. To calculate the fuzzy similarity relation $R_P$, which is used in fuzzy lower and upper approximations as mentioned in the equation \ref{RadzikowskaLower}, \ref{RadzikowskaUpper}, for the features subset $P \subseteq C$ the following equation may be taken.

\begin{equation}\label{fuzzySimilarityRelationFunction}
\mu_{R_P}(x,y)= \underset{f \in P}{\bigcap}\{\mu_{R_f}(x,y) \}
\end{equation}

Here, $\mu_{R_f}(x,y)$ denotes the similarity degree between object $x$ and $y$ with respect to feature $f$. Some examples of fuzzy similarity relations are given below:
\begin{equation}\label{eq1}
\mu_{R_f}(x,y) = 1- \frac{|f(x)-f(y)|}{|f_{max}- f_{min}|}
\end{equation}
\vspace{-1pt}
\begin{equation}\label{eq2}
\mu_{R_f}(x,y) = exp(-\frac{{(f(x)-f(y))}^2}{2\sigma^2})
\end{equation}

\noindent $\mu_{R_f}(x,y)$\\
\vspace{-5pt}
\begin{equation}\label{eq3}
\footnotesize
=max(min\bigg(\frac{(f(y)-(f(x)-\sigma_f))}{(f(x)-(f(x)-\sigma_f))},\frac{(f(x)+\sigma_f)-f(y))}{(f(x)+\sigma_f)-f(x))}\bigg),0)
\end{equation}
\normalsize
where $\sigma^2$ stands for the variance of feature $f$.\\
\begin{flushleft}
	\textbf{Upper approximation degree computation:}
\end{flushleft}

In Figure \ref{Pro_1}, the fuzzy rough upper approximation degree is computed as follows:

1. Computing fuzzy indiscernibility relation of conditional attributes using the Lukasiewicz t-norm and tolerance relation, as mentioned in section  \ref{FRUAsection}.

2. Computing fuzzy indiscernibility relation of decision attribute using its crisp relation.

3. Computing fuzzy upper approximation using the Lukasiewicz t-norm as per the equation \ref{RadzikowskaUpper}.

This fuzzy upper approximation degree can be used to select the samples from the negative samples set.

\subsubsection*{Data preprocessing for upper approximation degree computation:}

To reduce the dimension of feature vectors of the two datasets, we have utilized a dimensionality reduction method called incremental PCA. The feature dimension of a drug, target, and drug-target pair is 193, 1290, and  1483 for dataset1 and 881, 876, and 1757 for dataset2. 
To reduce the high computational cost of the fuzzy similarity computation (see equation  \ref{fuzzySimilarityRelationFunction}), we used incremental PCA to reduce feature dimension. This fuzzy similarity relation is further used to compute the upper or lower approximation. The computational complexity to compute the upper/lower approximation is $O(|N|^2 \times |D|)$ where $|N|$ is the size of the Universe and $|D|$ is the number of the decision classes. The computational complexity of the fuzzy similarity relation is $O(|N|^2 \times | C|)$ where $|C|$ is the number of attributes. For one single attribute, the similarity relation's computational complexity is $O(|N|^2 \times 1)$. For the attribute set $C$, there exist $|C|$ number of similarity relations in memory which incurs high computational cost. The situation goes, even more, worse for a high-dimensional dataset. To tackle this issue, we use incremental PCA which process the whole data by splitting it into mini-batches. Each batch can easily fit into the memory and is given as input to the incremental PCA at a time. Please note that the classical PCA and its variation (sparse-PCA, kernel-PCA) may also be applicable here, but this will results high computational cost, particularly for high dimensional data the algorithm may not be feasible in reality.


\LinesNotNumbered
\begin{algorithm}
	\small
	\DontPrintSemicolon
	\KwIn{$\textit{D}=\hspace{2pt} $feature matrix for the drug\\
		\hspace{38pt}$\textit{T}=\hspace{2pt} $feature matrix for the target\\
		
	}
	\KwOut{shared nearest neighbors represented by feature vectors}
	$k \gets$ Neighborhood size\\
	X $\gets$ D or T\\
	$n \gets$ sampleSize(X)\\
	distances = pairWiseDistance(X)\\
	sorted, indexes = sort(distances, ascendOrder)\\
	\For{$i\gets1$ \KwTo $n$ }{
		$\texttt{sharedNN = []}$\\
		\For{$j\gets1$ \KwTo $n$}{
			$\texttt{C = intersect(indexes(i,2:k+1),}$\\
			$ \texttt{indexes(j,2:k+1))}$\\
			$\texttt{sharedNN = sharedNN $\cup$ X(C)}$
		}
	}
	\caption{\small sharedNN}
	\label{sharedNNAlgo}
\end{algorithm}


\LinesNotNumbered
\begin{algorithm}
	
	\small
	\DontPrintSemicolon
	\KwIn{$\textit{DT}=\hspace{2pt} $ drug-target interaction matrix\\
		\hspace{42pt}$\textit{D}=\hspace{2pt} $feature matrix for the drug\\
		\hspace{42pt}$\textit{T}=\hspace{2pt} $feature matrix for the target\\
	}
	\KwOut{labeled TrainingDataSet}
	$P\gets \{ \hspace{2pt}\} $\hspace{6pt} \%$\hspace{2pt}\textit{P}=\hspace{2pt}$positive samples set\\
	$N\gets \{ \hspace{2pt}\} $\hspace{6pt} \%$\hspace{2pt}\textit{N}=\hspace{2pt}$negative samples set\\
	\For{$i\gets1$ \KwTo $m$ }{
		\For{$j\gets1$ \KwTo $n$}{
			
			\uIf{$DT(i,j)$=$1$}{
				$P \gets P\hspace{2pt}\cup$ concat(\(drugVec_i,targetVec_j\)) 
				
				/* $drugVec_i:i^{th}$ drug vector, $targetVec_j:j^{th}$ target vector */
				$tempD_i \gets sharedNN(drugVec_i)$\\
				$snnD_i \gets optimalKmedoidsCentroids (tempD_i)$ \\
				$tempT_j \gets sharedNN(targetVec_j)$\\
				$snnT_j \gets  optimalKmedoidsCentroids(tempT_j)$\\
				$N \gets N \hspace{2pt}\cup$ $cartesianProductPairConcat$(\(snnD_i,snnT_j\))\\
				
			}
		}
	}
	TrainingDataset $\gets P\cup N$\\
	
	\caption{Dataset Preparation}
	\label{DataSetPreparation}
\end{algorithm}

\begin{algorithm}[]
\small
\DontPrintSemicolon
\KwData{Imbalanced TrainingDataset $\mathcal{I}$ with $M$ samples $\{x_i,y_i\}$ where $i= 1$ to $M$ and $x_i$ is an d-dimensional vector in drug-target pair feature space and $y_i \in \{0,1\}$.
Assume $M_p$ and $M_q$ represent number of minority and majority class samples respectively, such that $M_p \leq M_q$ and $M_p + M_q = M$

}
\KwResult{BalancedTraingDataset}
Begin\\
function \textbf{upperAproxCalc}(decisionTable)\\
\hspace{12pt}    \textbf{begin}\\
\hspace{15pt}    $uDegree \rightarrow \{\}$ 
/* upper approximation degree vector */
\hspace{15pt}    $objCount \rightarrow sizeof(decisionTable)$ 
/* No. of object in decision table */

\hspace{15pt}    \For{$k\gets1$ \KwTo objCount }
{
$uDegree(k) \gets \mu_{\overline{R_{C}}Y}(decisionTable_k)$\\ here $C: conditional \hspace{4pt}attributes\hspace{4pt} set$ as per equation \ref{RadzikowskaUpper}

}
\hspace{12pt}    \textbf{end}\\
\vspace{7pt}
/* Split $M_p$ and $M_q$into $m$ and $n$ groups respectively */
$split(M_p) \rightarrow m\hspace{4pt} groups$ \\
$split(M_q) \rightarrow n \hspace{4pt}groups$\\

$totalNoGroupPair \gets m \times n$ 
/* total no. of group pairs between $m$ and $n$ */
$allGroupPairIndices \gets cartesianProduct(seq(1:m),seq(1:n))$ 
/* It holds $1$ to $m\times n$ indices where $i^{th}$ index holds $i^{th}$ pair */

\For{$i\gets1$ \KwTo totalNoGroupPair }
{
$allGroupPairIndices(i) \rightarrow (groupIndexOf_m,groupIndexOf_n)$ 
/* $groupIndexOf_m$, $groupIndexOf_n$: $m^{th}$ and $n^{th}$ group index no. from $m$ and $n$ groups respectively */
$decisionTable_i \rightarrow (P_{groupIndexOf_m} with \hspace{4pt}positive \hspace{4pt}label) \cup (N_{groupIndexOf_n} with \hspace{4pt}negative \hspace{4pt}label)$ 
/* $P_{groupIndexOf_m}$: set of positive samples taken from $groupIndexOf_m$, $N_{groupIndexOf_n}$:set of negative samples taken from $groupIndexOf_n$ */
$U_i \gets upperAproxCalc(decisionTable_i)$ 
$U_i$ holds upper approx. degree of all samples in $P_{groupIndexOf_m}$ and upper approx. degree of all samples in $N_{groupIndexOf_n}$ */

}

$\textbf{FRUA} : (\frac{1}{m}\sum (\hspace{4pt} upperApproxDegree \hspace{4pt}of \hspace{4pt}N_{groupIndexOf_n}\rvert$\\
$for \hspace{4pt}each \hspace{4pt} groupIndexOf_n \in seq(1:n)\hspace{4pt} and \hspace{4pt}\forall \hspace{4pt} groupIndexOf_m))$\\
\vspace{4pt}
\textbf{Sampling:}\\
$t_p$ and $t_q$ are the thresolds for $M_p$ and $M_q$\\
$Z \rightarrow \emptyset$\\

\For{$x \in M_q$}
{
   \If{$\textsc{FRUA}(x)\geq t_p$}
       {
       $M_p \gets M_p \cup x$

       }

   \If{$\textsc{FRUA}(x)\leq t_q$}
       {
       $Z \gets Z \cup x$

       }
}
BalancedTraingDataset= $ADASYN(M_p\cup Z)$

End
\caption{\small Average FRUA degree computation and sampling.}
\label{FRUAAlgoPseudoCode}
\end{algorithm}

\begin{figure}[!htb]
	\centering
	\includegraphics[scale=1.2]{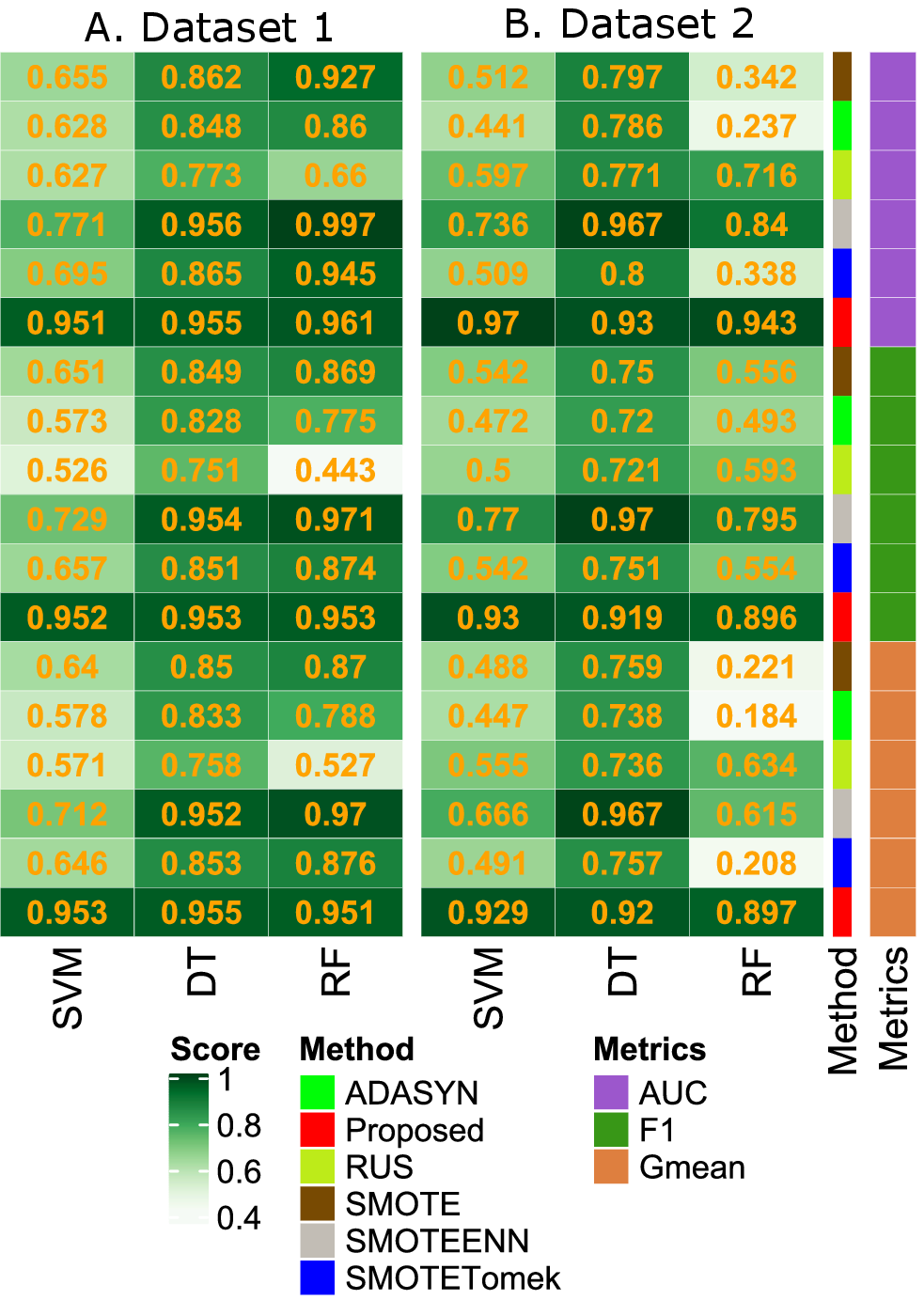} 
	\caption{Fig. (A) and (B) represents the performance on two datasets. The AUC, F1 and G-mean scores under the classification models of decision tree, random forest and support vector machine, respectively are demonstrated using various sampling methods.}
	\label{HeatMaps}	
	
\end{figure}

\begin{table}[]
\centering
\caption{Comparisons with the five state-of-the-arts methods}
\begin{tabular}{cccc}
\hline
\multicolumn{2}{c}{\multirow{2}{*}{Methods}}              & Dataset 1 & Dataset 2 \\ \cline{3-4} 
\multicolumn{2}{c}{}                                      & AUC       & AUC       \\ \hline
\multicolumn{2}{c}{RLS-avg, SVD}                          & 0.912     & 0.899     \\
\multicolumn{2}{c}{RLS-avg, PLS}                         & 0.915     & 0.918     \\
\multicolumn{2}{c}{RLS-avg, LapEig}                       & 0.909     & 0.916     \\
\multicolumn{2}{c}{RLS-kron, SVD}                         & 0.889     & 0.873     \\
\multicolumn{2}{c}{RLS-kron, PLS}                         & 0.899     & 0.913     \\
\multicolumn{2}{c}{RLS-kron, LapEig}                      & 0.889     & 0.874     \\
\multicolumn{2}{c}{EnsemDT, SVD}                          & 0.899     & 0.914     \\
\multicolumn{2}{c}{EnsemDT, PLS}                          & 0.902     & 0.898     \\
\multicolumn{2}{c}{EnsemDT, LapEig}                       & 0.901     & 0.914     \\
\multicolumn{2}{c}{EnsemKRR, SVD}                         & 0.942     & 0.931     \\
\multicolumn{2}{c}{EnsemKRR, PLS}                         & 0.941     & 0.930     \\
\multicolumn{2}{c}{EnsemKRR, LapEig}                      & 0.941     & 0.930     \\
\multicolumn{2}{c}{DeepPurpose}                           & 0.938     & 0.911     \\ \hline
\multicolumn{1}{l}{\multirow{4}{*}{Proposed}} & DT        & 0.955     & 0.930     \\
\multicolumn{1}{l}{}                          & RF        & 0.961     & 0.943     \\
\multicolumn{1}{l}{}                          & SVM       & 0.951     & 0.970      \\
\multicolumn{1}{l}{}                          & RUSBoost & 0.947     & 0.912     \\ \hline
\end{tabular}
\label{allOtherCompetingMethods}
\end{table}
\begin{table}
	\centering
	\caption{Drug-target interactions by proposed method}
	\scalebox{0.8}{
		\begin{tabular}{cccccc} 
			\hline
			\textbf{Drug}     & \textbf{Target}  & \textbf{FruaScore} & \textbf{Drug}   & \textbf{Target}    & \textbf{FruaScore}  \\ 
			\hline
			DB04094           & Q9Y296           & 0.933385           & \textbf{DB00839 } & \textbf{Q09428 } & \textbf{0.814468 }  \\
			DB03750           & P0CG47           & 0.933299           & \textbf{DB00476 } & \textbf{P28335 } & \textbf{0.810978 }  \\
			DB03988           & Q9Y296           & 0.933073           & \textbf{DB00450 } & \textbf{P35462 } & \textbf{0.806337 }  \\
			DB03320           & Q9Y296           & 0.932387           & \textbf{DB00776 } & \textbf{P35498 } & \textbf{0.804604 }  \\
			DB08242           & P0AEK4           & 0.932214           & \textbf{DB00929 } & \textbf{P43119 } & \textbf{0.803532 }  \\
			DB08137           & P0AEK4           & 0.932189           & \textbf{DB00433 } & \textbf{P35462 } & \textbf{0.802923 }  \\
			DB07153           & P16184           & 0.932128           & \textbf{DB00794 } & \textbf{Q14524 } & \textbf{0.799097 }  \\
			DB00992           & Q9Y296           & 0.932054           & \textbf{DB00917 } & \textbf{P21731 } & \textbf{0.798244 }  \\
			DB04789           & P16184           & 0.932053           & \textbf{DB01121 } & \textbf{Q14524 } & \textbf{0.795084 }  \\
			DB07000           & P0AEK4           & 0.932018           & \textbf{DB00645 } & \textbf{Q14524 } & \textbf{0.793230 }  \\
			DB04197           & Q9Y296           & 0.932002           & \textbf{DB00850 } & \textbf{P35367 } & \textbf{0.764447 }  \\
			DB07281           & P0AEK4           & 0.931912           & \textbf{DB04846 } & \textbf{P08913 } & \textbf{0.759809 }  \\
			DB03448           & P0A884           & 0.931780           & \textbf{DB00782 } & \textbf{P08172 } & \textbf{0.758948 }  \\
			DB04796           & P14867           & 0.931678           & \textbf{DB01365 } & \textbf{P08913 } & \textbf{0.751881 }  \\
			DB02456           & P0A884           & 0.931636           & \textbf{DB01121 } & \textbf{Q9NY46 } & \textbf{0.751538 }  \\
			DB04680           & P0CG29           & 0.931635           & \textbf{DB03719 } & \textbf{P30542 } & \textbf{0.747386 }  \\
			\textbf{DB01248 } & \textbf{P07437 } & \textbf{0.922451 } & \textbf{DB00670 } & \textbf{P08172 } & \textbf{0.745866 }  \\
			\textbf{DB00518 } & \textbf{P07437 } & \textbf{0.919137 } & \textbf{DB07954 } & \textbf{P30542 } & \textbf{0.744886 }  \\
			\textbf{DB00391 } & \textbf{P00915 } & \textbf{0.915100 } & \textbf{DB00794 } & \textbf{Q9Y5Y9 } & \textbf{0.730465 }  \\
			\textbf{DB01248 } & \textbf{Q13509 } & \textbf{0.914888 } & \textbf{DB00776 } & \textbf{Q9Y5Y9 } & \textbf{0.710952 }  \\
			\textbf{DB01248 } & \textbf{P68363 } & \textbf{0.911210 } & \textbf{DB00252 } & \textbf{Q9Y5Y9 } & \textbf{0.709006 }  \\
			\textbf{DB05294 } & \textbf{Q15303 } & \textbf{0.904014 } & \textbf{DB00999 } & \textbf{Q08460 } & \textbf{0.594489 }  \\
			\textbf{DB00361 } & \textbf{P68363 } & \textbf{0.897636 } & \textbf{DB01119 } & \textbf{Q08460 } & \textbf{0.589146 }  \\
			\textbf{DB01121 } & \textbf{P35499 } & \textbf{0.824893 } & \textbf{DB00356 } & \textbf{Q08460 } & \textbf{0.583733 }  \\
			\textbf{DB04846 } & \textbf{P07550 } & \textbf{0.816920 } & \textbf{DB03719 } & \textbf{P29274 } & \textbf{0.556650}  \\
			\hline
	\end{tabular}}
	\label{DTvalidWithCmap}
\end{table}
\begin{table}
	\centering
	\caption{Drug target interaction verification and new interaction by the proposed method}
	{\resizebox{.95\textwidth}{!}{
			\begin{tabular}{|m{1.1in}|m{0.08in}|m{2in}|m{2in}|} 
				\hline
				\multicolumn{2}{|c|}{} & \textbf{Correct prediction \textbf{of existing interactions}}  & \textbf{Novel Predicted interactions}  \\ 
				\hline
				\multirow{5}{*}{\begin{tabular}[c]{@{}c@{}} {}\\ \textbf{Target name:} \\ \textbf{Serine} \\ \textbf{hydroxymethyl}\\ \textbf{transferase,} \\ \textbf{ cytosolic} \end{tabular}} & \multirow{5}{*}{\rotatebox{90}{Drugs}} & Mimosine & Pyridostigmine \\ 
				\cline{3-4}
				&  & Pyridoxal phosphate & Willardiine \\ 
				\cline{3-4}
				&  & Glycine & acetamides \\ 
				\cline{3-4}
				&  & tetrahydrofolic acids& Betamipron \\ 
				\cline{3-4}
				&  & N-Pyridoxyl-Glycine-5-Monophosphate & Tyrosine \\ 
				\hline
				\multirow{5}{*}{\begin{tabular}[c]{@{}c@{}}\textbf{Target name:} \\\textbf{Monoamine}\\\textbf{oxidase} \end{tabular}} & \multirow{5}{*}{\rotatebox{90}{Drugs}} & Amphetamine & Diethylpropion \\ 
				\cline{3-4}
				&  & Phentermine & Ethinamate \\ 
				\cline{3-4}
				&  & Tranylcypromine & Alprenolol \\ 
				\cline{3-4}
				&  & Phenelzine & Phenylephrine  \\ 
				\cline{3-4}
				&  & Selegiline & Probenecid \\ 
				\hline
				\multirow{5}{*}{\begin{tabular}[c]{@{}c@{}} \textbf{Drug name:} \\ \textbf{alpha-D-} \\ \textbf{glucose } \\ \textbf{6-phosphate} \end{tabular}} & \multirow{5}{*}{\rotatebox{90}{Targets}} & Glucose-6-phosphateisomerase & Peptide deformylase \\ 
				\cline{3-4}
				&  & \begin{tabular}[c]{@{}c@{}} Glycogen phosphorylase, \\ muscle form \end{tabular} & \begin{tabular}[c]{@{}c@{}} Adenylate kinase \\ isoenzyme 1 \end{tabular}   \\ 
				\cline{3-4}
				&  & Aldose reductase & Adenosylhomocysteinase \\ 
				\cline{3-4}
				&  & \begin{tabular}[c]{@{}c@{}}Glutamine–fructose-6-phosphate \\ {aminotransferase [isomerizing]} \end{tabular} & Phosphoheptose isomerase \\ 
				\cline{3-4}
				&  & Hexokinase-1 & \begin{tabular}[c]{@{}c@{}}Low molecular weight2 \\ {tyrosine} \\ {protein phosphatase} \end{tabular}  \\ 
				\hline
				\multirow{5}{*}{\begin{tabular}[c]{@{}c@{}}\textbf{Drug name:}\\\textbf{Adenosine-5- } \\\textbf{Diphospho-} \\\textbf{ribose} \end{tabular}} & \multirow{5}{*}{\rotatebox{90}{Targets}} & \begin{tabular}[c]{@{}c@{}}MutT/nudix family protein \end{tabular} & \begin{tabular}[c]{@{}c@{}}Enoyl-[acyl-carrierprotein] \\ {reductase [NADH] FabI} \end{tabular} \\ 
				\cline{3-4}
				&  & \begin{tabular}[c]{@{}c@{}}p-hydroxy- \\{benzoate hydroxylase} \end{tabular}   & GDP-mannose6-dehydrogenase \\ 
				\cline{3-4}
				&  & \begin{tabular}[c]{@{}c@{}}Glyceraldehyde-3-\\ phosphate dehydrogenase \end{tabular} & \begin{tabular}[c]{@{}c@{}}RNA-directed\\ RNA polymerase \end{tabular} \\ 
				\cline{3-4}
				&  & Lactaldehyde reductase & \begin{tabular}[c]{@{}c@{}}Serine hydroxymethyl-\\ transferase \end{tabular} \\ 
				\cline{3-4}
				&  & Elongation factor 2 & Bifunctional protein BirA \\ 
				\hline
				\multicolumn{1}{c}{} & \multicolumn{1}{l}{} & \multicolumn{1}{c}{} & \multicolumn{1}{c}{}
			\end{tabular}
	}}{}
	\label{DTverificationPlusNew}
\end{table}

\section{Results and discussions}\label{MetResDis}
\subsection{Performance metrics}
This section explains the experimental results by using three metrics referred to as ROC-AUC scores, F1 scores, and Geometric Mean scores\cite{FAWCETT2006861}. The ROC-AUC provides a single score used to compare the models.
It ranges from $0$ to $1$ where $1$ indicates the perfect model and 0.5 represents a model having no prediction skill and the values less than 0.5 indicate that the prediction
skill is worse than no skill. The ROC-AUC performance evaluation is insensitive to highly imbalanced datasets. How well a model predicts the positive class and the negative class are represented by the sensitivity and specificity. The sensitivity and specificity together can be integrated into a single score called geometric mean is represented by \textit{sqrt(Sensitivity * Specificity)} where
the \textit{Sensitivity = TruePositive / (TruePositive + FalseNegative)} and \textit{Specificity=TrueNegative / (FalsePositive + TrueNegative)}.

The F1-score can be used to achieve a balance between Precision and Recall. It is also used where the class imbalance is present. All three scores are calculated using 5-fold cross-validation, and the average AUC, F1-score and G-mean score is computed. Note that the datasets $1$ and $2$ as mentioned in section \ref{MaterialsAndMethods} are used for prediction. 

\subsection{Proposed method vs some state-of-the-art sampling techniques}
The proposed method deals with imbalance classification problems for drug-target interaction prediction. We have compared it with the five state-of-the-art sampling techniques known as RUS, SMOTE,  ADASYN,  SMOTEENN, and SMOTETomek to deal with the imbalanced dataset.  Four classifiers, namely, decision tree(DT), random forest (RF), SVM, and RUSBoost are used to evaluate our proposed method’s performance. The ROC-AUC, F1, sand G-Mean scores of the proposed method, in Fig. \ref{HeatMaps}, are better than all the sampling methods. The RUS and SMOTE are performing poorly here for high-dimensional training data specified in \cite{Blagus2013}. ADASYN pays much attention to those samples of the minority class that are harder to learn. As our proposed method initially uses SNN, there may not be many samples that are harder to learn or the outliers.
For this reason, directly using ADASYN, unlike our proposed method, is not producing satisfactory results here. The Tomek’s link in SMOTETomek and edited nearest-neighbours in SMOTEENN is used to clean the noisy samples or marginal outliers in training data. The SMOTEENN and SMOTETomek are not performing well because there are no noisy samples or marginal outliers (due to shared nearest neighbours computation) in the training data.

\subsection{Comparisons with state-of-the-art methods}
We have compared the proposed method with five state-of-the-art methods, DeepPurpose\cite{Huang_2020}, RLS-avg (Regularized Least Squares-Average) \cite{Laarhoven2011GaussianIP} and RLS-kron (Regularized Least Squares-Kronecker product) \cite{10.1371/journal.pone.0066952}, EnsemDT \cite{Ezzat2017}, and EnsemKRR \cite{Ezzat2017}.
The DeepPurpose \cite{Huang_2020} is a deep learning-based method for drug-target interaction prediction. It is an encoder-decoder framework that uses eight encoders for a compound (drug) and seven encoders for an amino acid sequence (target). For this encoding, it uses deep neural networks, 1D convolutional neural networks, recurrent neural networks, transformer encoders, and message-passing neural networks.
The drug-target pairs, along with their fuzzy-rough upper approximation scores of our method, are compatible with the input data of the DeepPurpose model. The results in Table \ref{allOtherCompetingMethods}, show that the proposed method performs better than the DeepPurpose for ROC-AUC score with the same data.
For each of the remaining methods, we have utilized three different dimensionality reduction techniques, namely Singular Value Decomposition(SVD), Partial Least Squares (PLS), and Laplacian Eigenmaps (LapEig) for the preparation of training data. The results in Table \ref{allOtherCompetingMethods}, show that our proposed method has satisfactory ROC-AUC results (0.955, 0.961, 0.951, 0.947 for dataset-1 and 0.930, 0.943, 0.970 and 0.912 for dataset 2 using DT, RF, SVM and RUSBoost classifier respectively. 

We have only provided the ROC-AUC scores of all these competing methods due to unavailability of the F1 and G-Mean scores in \cite{Ezzat2017}. The parameters of RLS-avg, RLS-kron, EnsemDT, and EnsemKRR are set to the default values as specified in \cite{Laarhoven2011GaussianIP}, \cite{10.1371/journal.pone.0066952}, and \cite{Ezzat2017}
 
 \subsection{Tuning of hyperparameters}
 
 The proposed method performs grid search-based hyperparameter tuning for computing ROC-AUC, F1, and G-Mean scores. For the DT classifier, we have observed that the best ROC-AUC, F1, and G-Mean scores are obtained using the hyperparameters combination is \textit{criterion: gini, maxDepth: 9, minSamplesLeaf: 1, minSamplesSplit: 6} for dataset 1. For dataset 2, the best ROC-AUC, F1, and G-Mean scores have been achieved by \textit{criterion: gini, maxDepth: 9, minSamplesLeaf: 1, minSamplesSplit: 4}. 
 In the case of RF classifier, for dataset 1 and dataset 2, the best hyperparameters combination is determined as \textit{criterion : gini, maxDepth: 20, minSamplesLeaf: 3, minSamplesSplit: 8, nEstimators: 200} for ROC-AUC scores of 0.961 and 0.943, respectively.
 Fig. \ref{HyperParameters} (A) and (B) demonstrate the variation of the AUC score of the decision tree with respect to only two hyperparameters called \textit{tree\_depth} and \textit{max\_feature}. In Fig. \ref{HyperParameters} (C), a heatmap is shown only for hyperparameters (\textit{n\_estimators}, \textit{max\_depths}) for the random forest model. The maximum depth of the tree is decided as nodes are expanded until all leaves are pure or until all leaves contain less than $minSamplesSplit$ samples. The number of features for both the RF and DT is equal to maxFeatures = sqrt(nFeatures).
The best hyperparameters combinations in SVM for dataset 1 are determined as \textit{kernel: RBF, C: 10.0, gamma: 0.1}. As for dataset 2, the best ROC-AUC, F1, and G-Mean scores are 0.97, 0.93, and 0.929 achieved using kernel: RBF, C: 1.0, gamma: 0.1. Fig. 4 (D) represents the ROC-AUC scores with two hyperparameters (C, gamma) for dataset 2.

To prepare negative drug-target pairs, the number of nearest neighbours is $11$, which is later used to compute the shared nearest neighbours. We observed that for 11 nearest neighbours, the shared nearest neighbours computation step determines the number of drugs and targets that have a good balance between the number of samples and feature dimension.


\subsection{Feature selection and comparisons}
\begin{figure}[!htb]
	\centering
	\includegraphics[scale=0.35]{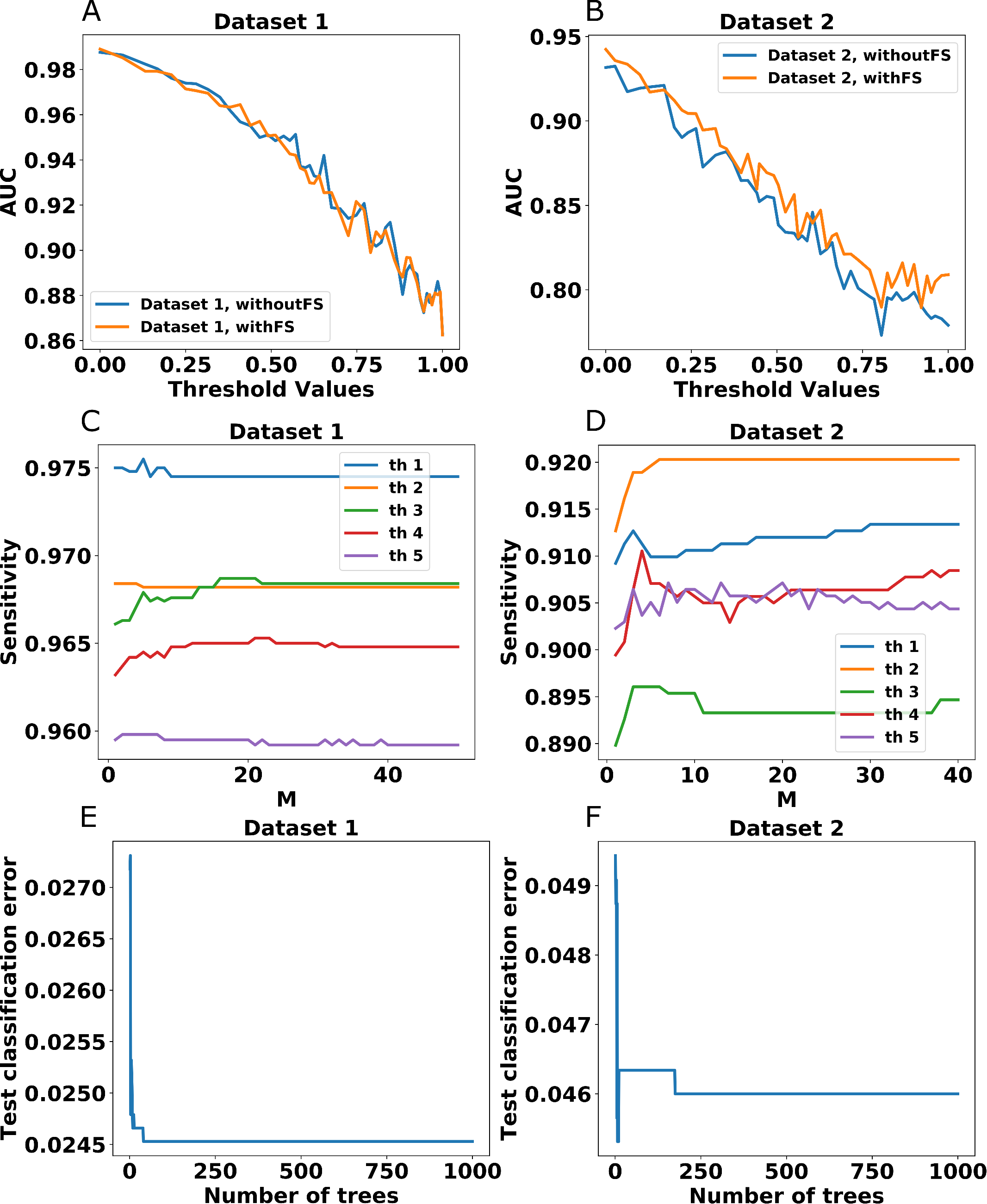}
	\caption{Fig. (A) and (B) represent Threshold vs AUC graph for dataset 1 and dataset 2 using feature selection and without feature selection respectively. (C) and (D) represent M vs Sensitivity plots for both datasets using five thresholds. (E) and (F) represent classification errors for both dataset 1 and dataset 2, respectively using one threshold.}
	\label{allFigures}	
\end{figure}

\begin{figure}[!htb]
	\centering
	\includegraphics[scale=0.35]{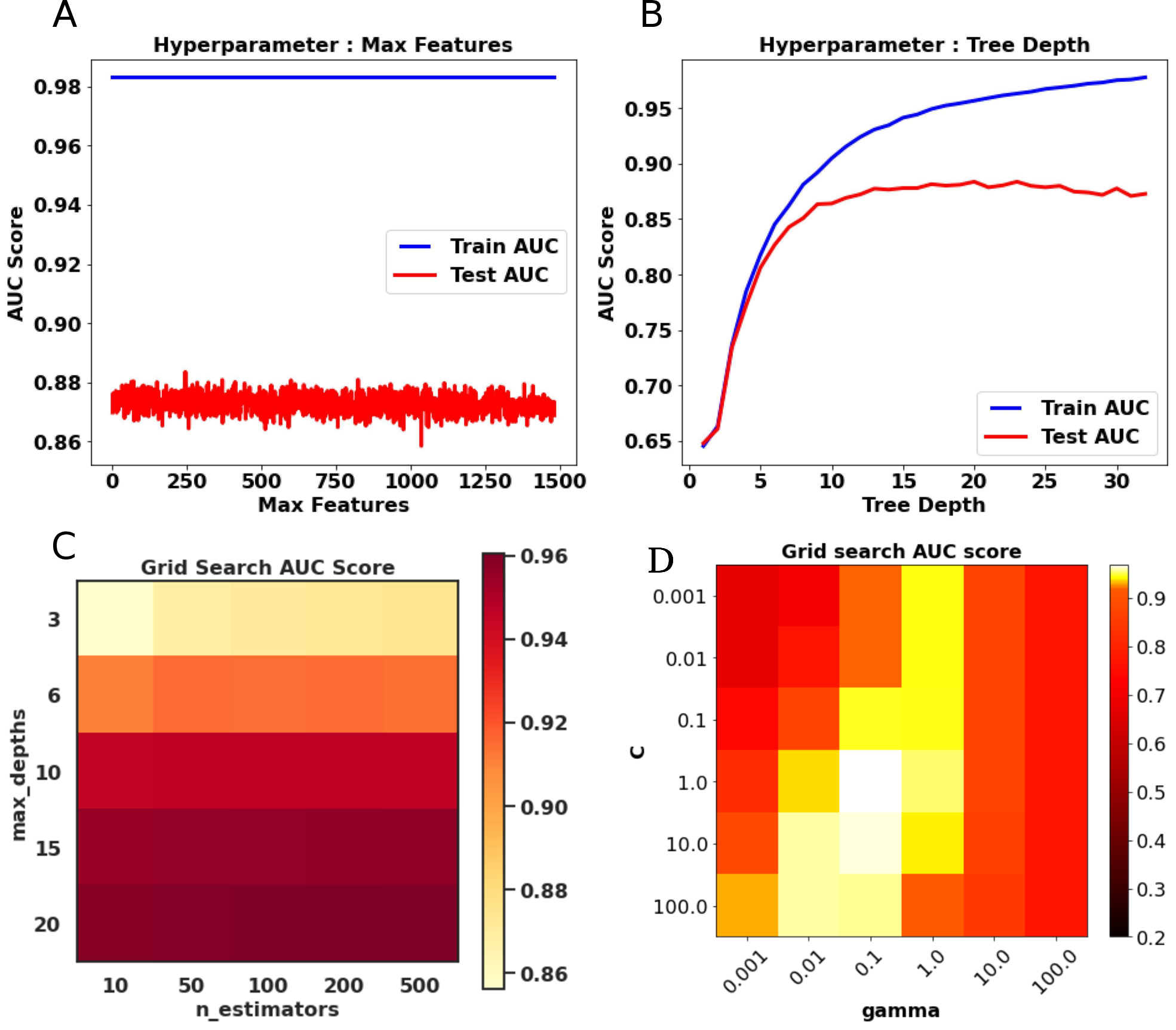}
	\caption{Fig. (A) and (B) represent the hyperparameters of decision tree called max feature and tree depth vs AUC graph for dataset $1$, respectively. In (C), the hyperparameters of random forest along with the AUC scores are shown in the heatmap. Fig. (D) represents one heatmap for AUC scores of SVM for two hyperparameters called C and gamma.}
	\label{HyperParameters}	
	
\end{figure}

In Fig. \ref{allFigures} (A) and (B), the prediction scores in terms of ROC-AUC values have been shown for both datasets considering feature selection and no feature selection.  In our method, after SNN computation followed by k-medoids clustering, we have computed a fuzzy rough upper approximation score (grade membership degree) as the strength of the interaction between a drug and a target for each of the unannotated pairs. Based on different threshold cut-off values of this score, we divided all the unannotated drug-target pairs into positive and negative classes. Negative samples detected from the unannotated pairs via fuzzy rough upper approximation score and the initially obtained annotated positive samples constitute the input data for RUSBoostClassifier. 
For different threshold cut-off values of fuzzy rough upper approximation scores, the RUSBoostClassifier produces the  Fig. \ref{allFigures} (A) and (B). In these experiments, we used the holdout strategy for training with the training and testing ratio of 70:30. Table \ref{allOtherCompetingMethods}, the ROC-AUC scores of RUSBoostClassifier for one threshold cut-off value, for dataset 1 and dataset 2, are obtained by executing hyperparameters tuning using grid search. The best hyperparameters are determined as $nEstimators : 500, learningRate: 1.0$ which produces 0.9477 and 0.912 for ROC-AUC for dataset 1 and dataset 2. 
The RUSBoostClassifier is used here because it mitigates the class imbalance problem during learning by the random under-sampling of the samples at each iteration of boosting.
For feature selection, the features importance scores have been computed using XGBoost and random forest. These two feature importance computation methods split the positive and negative samples into many groups, where the number of positive and, negative samples in each group is approximately equal. All the positive and negative group pairs were individually taken by the XGBoost and random forest classifiers for computing the feature importance. Finally, average feature importance scores are computed and top $100$ features are taken for prediction. The average execution time, without feature selection, over $50$ thresholds for dataset 1 and dataset 2 are $617.66$ sec., and $232.07$ sec., respectively. When feature selection is considered, the average execution time, over $50$ thresholds, for dataset 1 and dataset 2 are $232.07$ sec., and $77.61$ sec., respectively.

\subsection{Sensitivity vs number of base learners and classification errors}
In Fig. \ref{allFigures} (C) and (D), two plots represent the M vs Sensitivity graph for both datasets where M represents the number of base learner that is ranging from $1$ to $50$. This experiment is carried out for a few threshold values. For each threshold, the variation of the ROC-AUC is minimal. The classification error indicates the proportion of samples that the classifier misclassified are also reported in  Fig. \ref{allFigures} (E) and (F).

\subsection{Drug-target interaction of the proposed method}
\begin{figure}[!htb]
	\centering
	\includegraphics[scale=0.45]{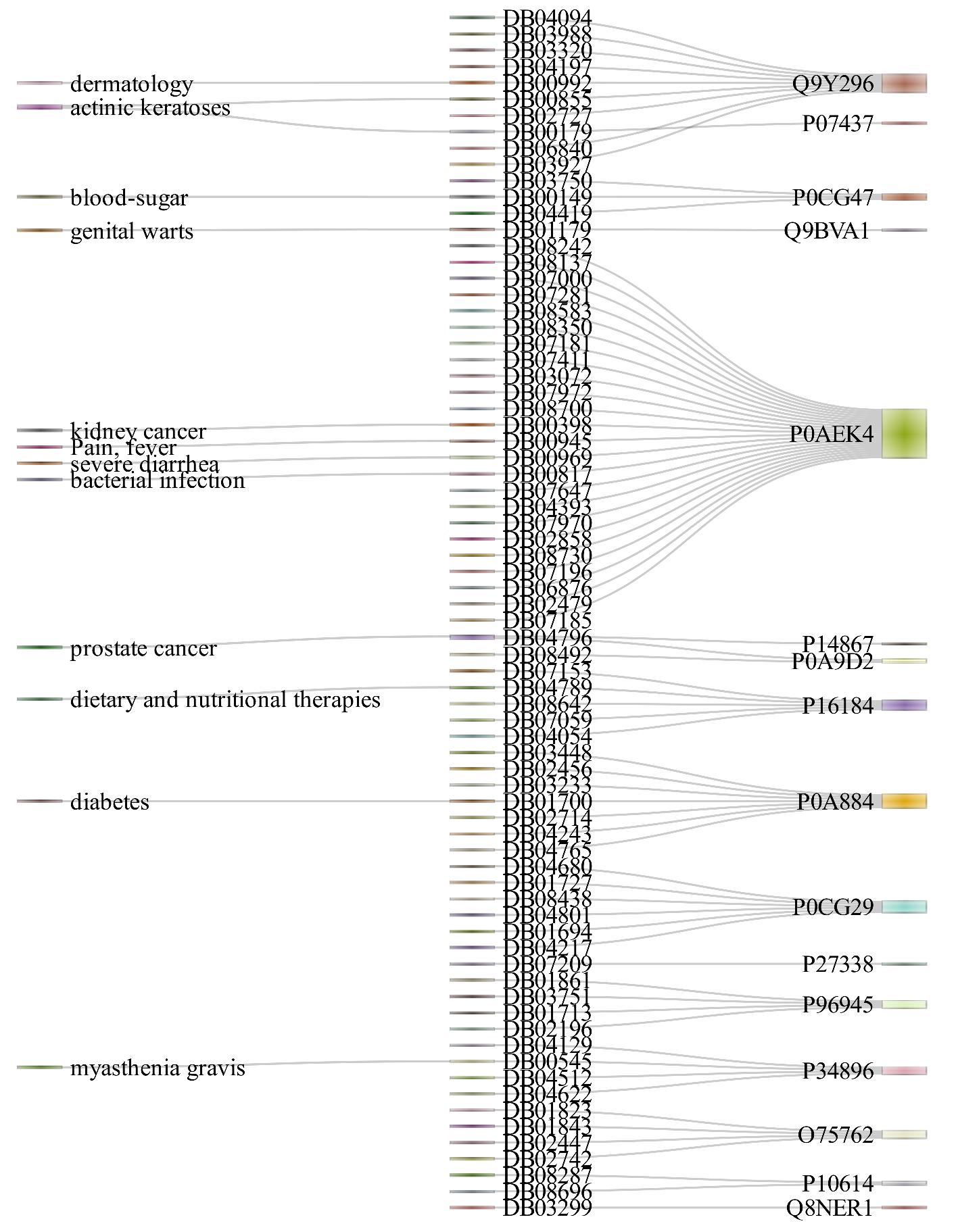}
	\caption{Some drug-target interactions with treatment areas of the drugs.}
	\label{SanKeyPlot}
\end{figure}

In Table \ref{DTverificationPlusNew}, some existing and predicted drug-target interactions have been provided. To test the efficacy of the proposed method, we have omitted several known interactions from training data. Then, we have trained our model with the remaining data and verified our prediction results. We have observed that our prediction model has even successfully recovered (predicted) those omitted known interactions. Seven drugs for the target \textit{Serine hydroxymethyltransferase, cytosolic} are predicted correctly, and among them, five are listed in Table \ref{DTverificationPlusNew}. For the same target, we predicted five additional interactions with drugs. Similarly, we have displayed results of some correctly predicted and novel drug-target interactions in this table.
In Fig. \ref{SanKeyPlot}, some drug-target interactions have been shown, along with some interactions between the treatment areas and drugs.


\subsection{Drug-target interaction validation}

To verify our drug-target interaction prediction results, we have used the Connectivity Map (Cmap) \cite{Iorio14621} prediction results provided by the Broad Institute. The drug name and target name in the Drugbank dataset have different representations in Cmap.  Therefore, we have performed the conversion between Drugbank ID and Cmap using the webchem R package \cite{webChemCitation}.  This R package retrieves the chemical information from the web using a suite of 14 web services.

Our prediction results of drug-target pairs for Drugbank dataset are utilized in the webchem packages, which only fetches information from the Wikidata. Due to lack of information in the suite of web services, except the Wikidata, as provided by webchem R package, we have not obtained complete matching between our prediction and Cmap predictions. In Table \ref{DTvalidWithCmap}, a list of $50$  drug-target interaction pairs is shown that has been predicted by our method. Thirty-four interaction pairs which are also available in the Cmap predicted database is marked in bold face.

We have also observed that most of predicted drug-target interaction pairs  e.g. {(DB01248, P07437), (DB04846, P07550), (DB00839, Q09428), (DB00450, P35462), (DB00776, Q9Y5Y9), (DB00776, P35498)} shown in Table \ref{DTvalidWithCmap},  are also reported in \cite{MATESANZ2008573}, \cite{Yao2008EffectsOT}, \cite{ASANO19992289}, \cite{PMID:20358234}, \cite{PMID:20502724} and \cite{PMID:20358234}.

\section{Conclusion}\label{conclu}
In this article, a novel computational approach for drug-target interaction prediction is presented utilizing existing drug-target data. There are two critical issues in this domain: a massive amount of drugs and targets creating a vast search space and highly imbalanced drug-target interactions dataset as there is a tiny number of drug-target interactions unveiled so far.
Thus, the size of the negative samples is much larger than the size of the positive samples.

Here, we have used shared nearest neighbours rather than taking a fixed number of nearest neighbours as it is more effective in the higher dimensional dataset. The reason behind this is, typically, the size of the overlapped items within the neighbourhoods of a pair of drugs (or targets) inside the same cluster is substantially larger than the neighbourhoods of a pair of drugs (or targets) belonging to different clusters. 
Moreover, to tackle the curse of the imbalanced dataset, these shared nearest neighbours are further grouped by k-medoids. The representative centroids of k-medoids for the drug and target are then considered new possible drug-target interaction pairs for each known drug-target pair. Additionally, to deal with imbalanced dataset further, we have computed the degree of fuzzy-rough upper approximation of all the possible interaction pairs in the negative samples to perform undersampling. After that, selecting a threshold of the computed degrees, the size of the negative and positive samples sets are balanced. This upper approximation degree-based undersampling of the negative samples causes improvement in the prediction scores. Computation of degree in the fuzzy-rough upper approximation is challenging as interaction pairs' dimension is exceptionally high. The execution time of this fuzzy-rough upper approximation degree is directly proportional to the number of features. Therefore, further investigation on fuzzy-rough set based feature selection followed by fuzzy-rough upper approximation computation may improve the prediction score. Instead of using a single threshold for undersampling, multiple threshold-based undersampling may be investigated for tackling the curse of imbalanced datasets. Moreover, the positive samples' oversampling to balance with the number of negative samples may be explored to improve the prediction score. We believe that DTI-SNNFRA may be a promising framework for drug-target interaction prediction.

\nolinenumbers

%
%
%
%
%
%
%

\bibliographystyle{plos2015}

\end{document}